\documentclass[a4paper,twoside]{article}

\usepackage{epsfig}
\usepackage{subfigure}
\usepackage{calc}
\usepackage{amssymb}
\usepackage{amstext}
\usepackage{amsmath}
\usepackage{amsthm}
\usepackage{multicol}
\usepackage{pslatex}
\usepackage{apalike}
\usepackage{SCITEPRESS}     

\begin{document}

\title{Practical License Plate Recognition in Unconstrained Surveillance Systems with Adversarial Super-Resolution}

\author{\authorname{Younkwan Lee, Jiwon Jun, Yoojin Hong, Moongu Jeon}
\affiliation{Machine Learning and Vision Laboratory, Gwangju Institute of Science and Technology, Gwangju, South Korea}
\email{\tt\small \{brightyoun, battery94, yoojinhong, mgjeon\}@gist.ac.kr}
}


\abstract{Although most current license plate (LP) recognition applications have been significantly advanced, they are still limited to ideal environments where training data are carefully annotated with constrained scenes. In this paper, we propose a novel license plate recognition method to handle unconstrained real world traffic scenes. To overcome these difficulties, we use adversarial super-resolution (SR), and one-stage character segmentation and recognition. Combined with a deep convolutional network based on VGG-net, our method provides simple but reasonable training procedure. Moreover, we introduce GIST-LP, a challenging LP dataset where image samples are effectively collected from unconstrained surveillance scenes. Experimental results on AOLP and GIST-LP dataset illustrate that our method, without any scene-specific adaptation, outperforms current LP recognition approaches in accuracy and provides visual enhancement in our SR results that are easier to understand than original data. }

\onecolumn \maketitle \normalsize \vfill

\section{\uppercase{Introduction}}
\label{sec:introduction}

    License plate recognition (LPR) is a fundamental and essential process of identifying vehicles and can be extended to a variety of real-world applications. LPR methods have been widely studied over the last decade, and are especially of big interest in intelligent transport systems (ITS) applications such as access control \cite{chinomi2008prisurv}, road traffic monitoring \cite{noh2016adaptive,pu2013t,song2016online,lee2017automatic,yoon2018online} and traffic law enforcement \cite{zhang2011data}. Since all license plate recognition methods always deal with the letters and numbers in images, they are closely related to image classification \cite{simonyan2014very,russakovsky2015imagenet} and text localization \cite{anagnostopoulos2006license}. 
    
    \begin{figure}[t]
    \begin{center}
       \includegraphics[width=1.0\linewidth]{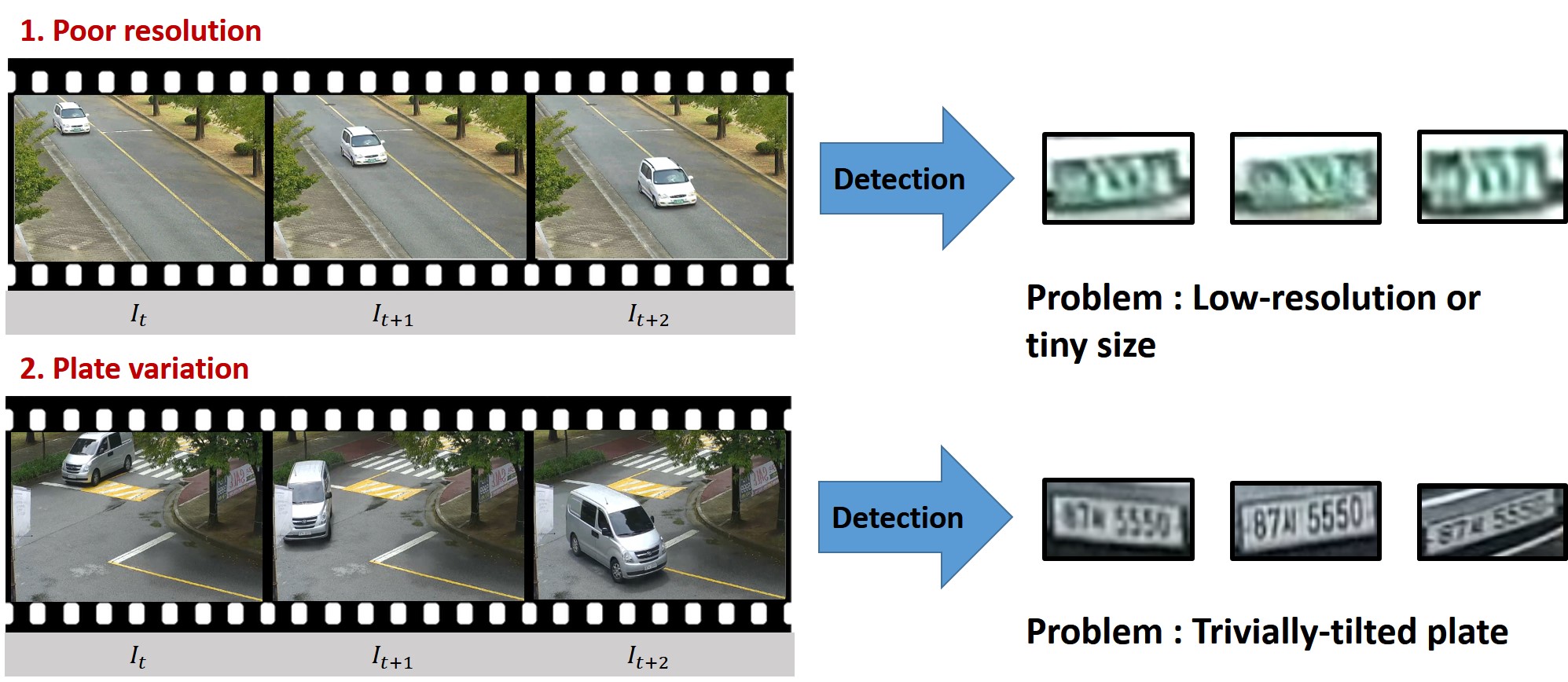}
    \end{center}
       \caption{Example in GIST-LP dataset. Poor-resolution and plate variation are common challenging issues on license plate recognition problem.}
    \label{fig:long}
    \label{fig:onecol}
    \end{figure}

    Conventional LPR methods typically include two stages: character localization and character recognition. Those methods are widely designed for unrealistically most constrained scenarios: a high-quality resolution and an unrotated frontal or rear image. However, unlike the ideal situation, many traffic surveillance cameras scattered around the world are operating in a number of unconstrained scenarios: they produce poor-resolution images and tilted license plates as shown in Figure 1. Although considerable progress of computer vision technology has been made, existing methods may fail to recognize license plates in such an environment without considering any unconstrained conditions. As a consequence, we find its limitations in three aspects: first, many license plate samples only constitute incomplete text search space; second, the projection angle of the sample is tilted with respect to the image plane at an angle of up to 30 degrees, interfering character exploitation; third, bad text localization often results in erroneous outputs. 
  
    Based on this finding, we propose a novel deep convolutional neural network based method for better LPR.
    
    \textbf{Adversarial Super-Resolution} We suggest an adversarial super-resolution (SR) method including a generator and a discriminator networks over an image area. Modern SR method \cite{dong2014learning} commonly targets the pixel-wise average as optimization goal, minimizing the mean squared error (MSE) between the super-resolved image and the ground truth, which leads to the smoothing effect, especially across text. Instead, we follow \cite{ledig2017photo}'s generator network, which solves minimax game as optimization goal, avoiding a smoothing effect, which provide a sharpening effect. Combined with SR in generator, we introduce a new loss function that encourages the discriminator to count characters and distinguish whether SR or high-resolution(HR) sample concurrently. Character counting results from the discriminator network  help improve character recognition performance in one-stage recognition module as a conditional term.
    
    \textbf{Reconstruction Auto-Encoder} We always reconstruct the samples to straighten when the horizontally or vertically tilted license plate is projected onto the image plane. To address this issue, we utilize the convolutional auto-encoder network with the objective function as the difference between the tilted image and the straightened image. By doing so, it serves as a preprocessing for correct character exploitation. 
    
    \textbf{One-Stage Recognition} We do use the commonly used character segmentation and localization process. Instead, we propose a unified character localization and recognition approach as one-stage. One-stage recognition is not only more intuitive, but also more accurate than segmentation that requires precise an estimate of each pixel's class. Our One-stage method divides the input image into a 1*S grid, and detects LP at three different scales which includes a conditional term. The result of our character localization using each grid cell is naturally unified with character classification.

    In summary, our key contributions are:
    \begin{itemize}
      \item[$\bullet$] We show that adversarial SR module and AE based reconstruction module in the real world for unconstrained surveillance cameras can improve the recognition performance greatly by (2.57\% (AOLP) and 8.06\% (GIST-LP)) compared with the state-of-the-art methods.
      \item[$\bullet$] The One-stage method combined with the conditional term, instead of the two-stage method (character detection and classification), reduced the localization and classification error.
      \item[$\bullet$] We collected a dataset of challenging license plate samples from unconstrained conditions accompanied by the text annotations (1,800 samples, 50 different license plates).
    \end{itemize}

    \begin{figure*}[t]
    \begin{center}
       \includegraphics[width=1.0\linewidth]{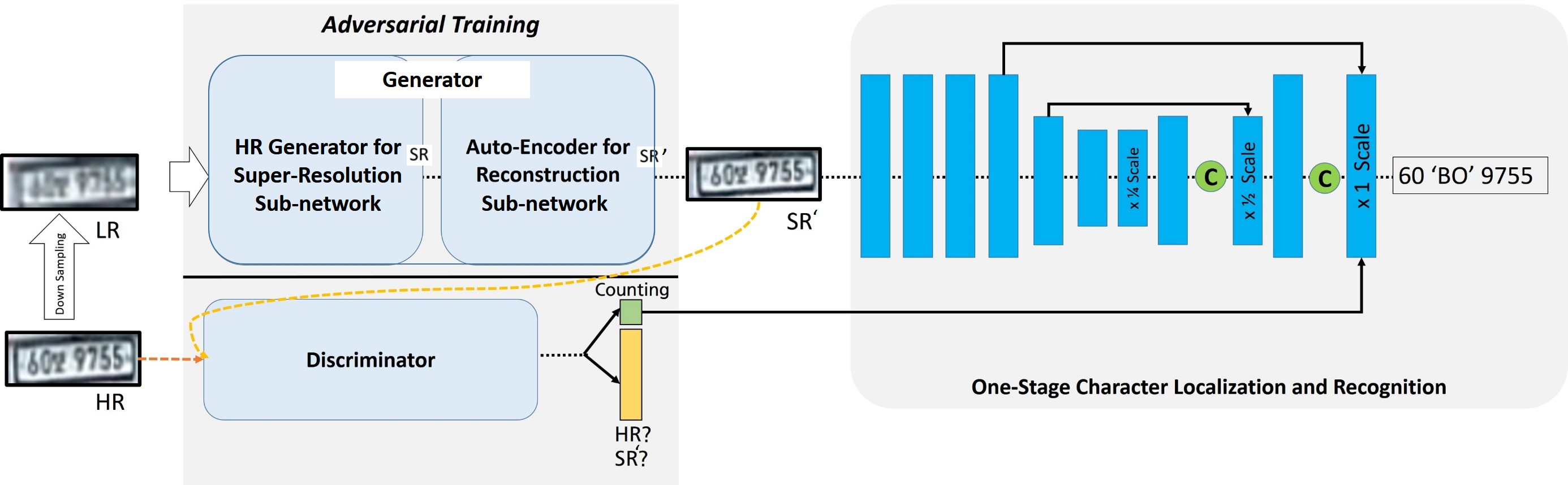}
    \end{center}
       \caption{The proposed license plate recognition pipeline.}
    \label{fig:long}
    \label{fig:onecol}
    \end{figure*}

\section{\uppercase{Related Work}}

\subsection{License Plate Recognition}
Traditionally, numerous LPR methods proposed consists of the two stages: semantic segmentation of the exact character region and recognition of the characters. The related methods generally utilize discriminate features, such as edge, color, shape and texture but does not show good results. Edge-based methods \cite{kim2000learning,zhang2006learning,wang2003detection,kim2000learning,zhang2006learning} and geometrical features \cite{wang2003detection} assume the presence of characters in the license plate. Many color-based methods \cite{shi2005automatic,chen2009automatic} usually use the combination of the license plate and the characters. 

However, since the two-stage methods are not only slow to run, but also take more time to converge for optimized training due to the double networks, one-stage pipeline based methods, segmentation-free approach \cite{zherzdev2018lprnet,cheang2017segmentation,li2016reading,wangadversarial}, including segmentation and recognition at once, are proposed. Most segmentation-free models take advantage of deeply learned features which outperforms traditional methods on the task of classification by deep convolutional neural networks (DCNN) \cite{simonyan2014very,he2016deep} and data-driven approaches \cite{russakovsky2015imagenet}. The core underlying assumption of these methods extracts features directly without sliding window for LPR. As examples of these models, Sergey \textit{et al.} \cite{zherzdev2018lprnet} adopted a lightweight convolutional neural network to learn end-to-end way. In another work that use RNN module, Teik Koon \textit{et al.} \cite{cheang2017segmentation} proposed CNN-RNN unification model that feed the entire image as input. It is assumed that the context of the entire image is further evaluated for exact classification than the sliding window approaches being. Also, Hui \textit{et al.} \cite{li2016reading} utilized a cascade framework using DCNN and LSTM and Xinlong \textit{et al.} \cite{wangadversarial} proposed DCNN and a bidirectional LSTM to use sequence labeling.

\subsection{Adversarial Learning}
The generative adversarial network (GAN) ~\cite{goodfellow2014generative,radford2015unsupervised,radford2015unsupervised} is an amazing solution for training deep neural network of generative models, which aim to learn the probability distributions of the input data. Originally, GAN is suggested to yield the more realistic-fake images~\cite{frid2018gan}, but recent researches show that this adversarial technique can be utilized to produce the specific training algorithms. e.g,. generative focused tasks; super-resolution ~\cite{nguyen2016plug,ledig2017photo,lee2018accurate}, style transfer ~\cite{zhu2017unpaired,li2017demystifying}, natural-language processing ~\cite{rajeswar2017adversarial} and discriminative focused tasks; human pose estimation ~\cite{chou2017self,peng2018jointly}.

\section{\uppercase{Proposed Method}}

In this section, we describe the details of the proposed end-to-end pipeline for LPR. The schematics of the method is illustrated in Figure 2. We first  introduce the adversarial network to super-resolve the input image, and reconstruct its output. Then, the details of the proposed one-stage character recognition network are presented for recognizing characters on the license plate and locating individual text regions without character segmentation. Finally, we describe a training process to find optimal parameters of our model.

    \begin{figure*}[t]
    \begin{center}
       \includegraphics[width=1.0\linewidth]{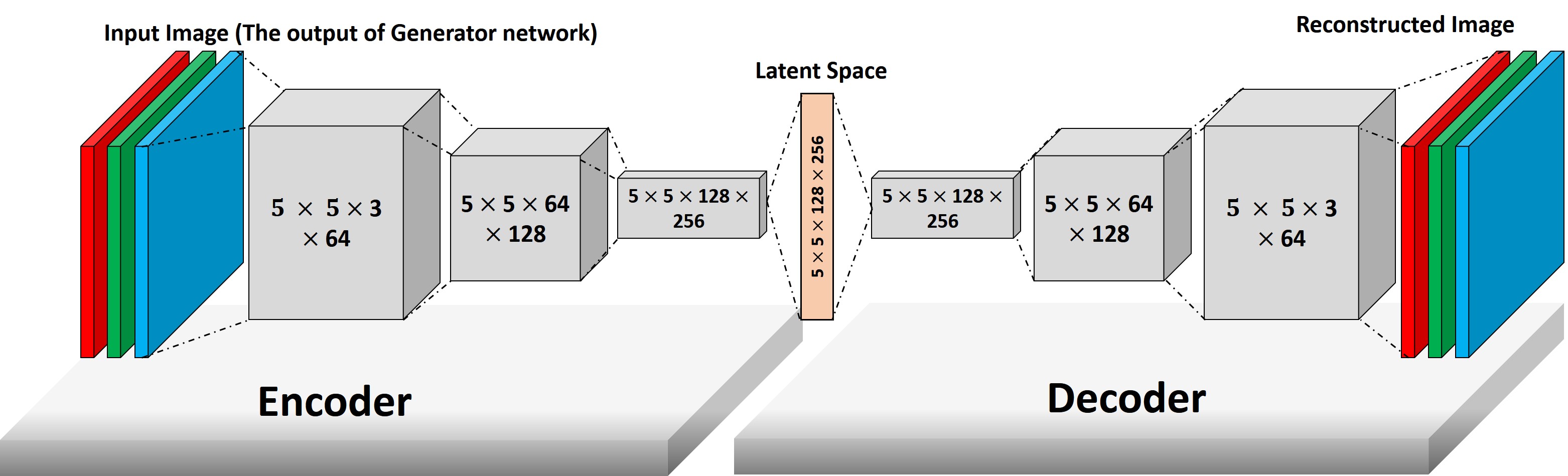}
    \end{center}
       \caption{The proposed Auto-Encoder based reconstruction sub-network structure.}
    \label{fig:long}
    \label{fig:onecol}
    \end{figure*}
    
\subsection{Adversarial Network Architecture}
    Adversarial learning techniques have been widely used in many tasks \cite{frid2018gan,zhu2017unpaired,rajeswar2017adversarial,chou2017self}, providing boosted performance through adversarial data or features. In vanilla GAN \cite{goodfellow2014generative}, a minimax game is trained by alternately updating a generator sub-network $G$ and a discriminator sub-network $D$ simultaneously. The value function of the generator $G$ and the discriminator $D$ is defined as:
        \begin{equation}
        \begin{split}
            \min_{\theta_G} \max_{\theta_D} V(D,G) = &\mathbb{E}_{x\sim p_{real}~(x)}[log D(x)] \\
            & + \mathbb{E}_{z\sim p_{fake}(z)}[log(1-D(G(z)))]
        \end{split}
        \end{equation}
    where $p_{real}$ is the real data distribution observation from $x$ and $p_{fake}$ is the fake data distribution observation from a random distribution $z$. These sub-networks have conflicting goals to minimize their own cost and maximize the opposite's cost. Therefore, the conclusion to play the minimax game can be that the probability distribution ($p_{fake}$) generated by the generator $G$ exactly matches the data distribution ($p_{real}$). After all, the discriminator $D$ will not be able to distinguish between sampling distribution from the generator $G$ and real data distribution. At this time, for the fixed generator, the optimal discriminator function is as follows:
        \begin{equation}
            D_{G}^*(x) = \frac{p_{real}(x)}{p_{real}(x) + p_{fake}(x)}.
        \end{equation}
    
    In a similar way, we modified the minimax value function in the vanilla GAN for solving SR so that the generator $G$ consisting of a HR generator $G_{SR}$ and a reconstruction network $G_{recon}$ creates an HR image from LP image, while the discriminator $D$ trains to distinguish the HR fake image obtained by the generator from the actual LR image. This adversarial SR process can be defined as follows:
    	\begin{equation}
    	\begin{split}
        	&\min_{\theta _G} \max_{\theta _D} V(D,G) = \mathbb{E}_{I^{HR} \sim p_{train}(I^{HR})}[\log D_{\theta _D}(I^{HR})] \\
        	& + \mathbb{E}_{I^{LR} \sim p_{G}(I^{LR})}[\log (1 - D_{\theta _D}(G_{\theta _G}I^{LR}))],
        \end{split}
    	\end{equation}
     where $I^{HR}$ is the high-resolution image, $I^{LR}$ is the low-resolution image, $\theta _G$ and $\theta _D$ denote the parameters trained by a feed-forward CNN $G_{\theta _G}$ and $D_{\theta _D}$ respectively. 
     
     \textbf{Generator Network.} Different from \cite{goodfellow2014generative}, our generator network is composed of two sub-networks: (1) HR Generator $G_{SR}$ and (2) Convolutional Auto-encoder for reconstruction $G_{recon}$ as shown in Figure 2. The former is a series of convolutional layers and fractionally-strided convolution layers (\textit{i.e. upsample layer}) inspired by \cite{ledig2017photo}. We use two upsample layers(2 times upsampling) as proposed by Radford et al. \cite{radford2015unsupervised}, and acquire a 4 times enhanced image image from them. 
     
     In addition to its network, we include a reconstruction sub-network for the refinement task of image with enhanced resolution. Given the output of 4 times super-resolved image, our proposed network aims at discovering that it corrects slightly distorted image through denoising learning manner. Basically, we employ a convolutional neural network (CNN) as encoder and decoder, as shown in Figure 3. Although both encoder and decoder consist of the same number of convolutional layers, the former adds MaxPooling2D layers for spatial down-sampling, while the latter adds UpSampling2D layers, with the BatchNormalization \cite{ioffe2015batch}. 
     
     \textbf{Discriminator Network.} Figure 2 shows the architecture of the discriminator network and its output components. Inspired by VGG19 \cite{simonyan2014very}, we follow the same network structure. To discriminate exact object regions, we design all the fully-connected layers to split into two parallel branches to obtain two outputs: (1) how many characters are in the image as counting result $f_{count}$ and (2) the HR \textit{vs.} SR $f_{GAN}$. 
     
\subsection{Character Recognition Network Architecture}
     In this section, we describe the details of the proposed character recognition approach where localization and recognition are integrated into one-stage. We employ YOLO v3 \cite{redmon2018yolov3} as our detection network. To achieve scale-invariance, it detects characters at three scales, which are given by diminished dimensions of the image by 32, 16 and 8 each other, without the MaxPooling2D layer. Unlike previous model \cite{redmon2017yolo9000}, this allows better detection performance of small size character, which is optimized for character on a license plate that is mostly expressed in small size localization and recognition with residual skip connections. 
     
     The shape of detection kernel denoted as 1 $\times$ 1 $\times$ ($B$ $\times$ (5 + $C$)), where $B$ is the number of bounding boxes, $5$ is the sum of the four attributs of bounding boxes (coordinates ($x$, $y$), width and height) and one object confidence score and $C$ is the number of classes. In our method, we define the detection kernel size as $B = 3$ and $C$ is 66 (10 numbers (0-9), 26 English letters and 30 Korean letters), result in 1 $\times$ 1 $\times$ 213.  
     
     Furthermore, we add the counting information output $f_{count}$ from the discriminator as a conditional term in our character recognition model. The last layer of recognition model has the previous layer's output and $f_{count}$ as inputs. We demonstrate that our recognition model can be extended to the sophisticated model where it can accurately count and localize any character in any input. These are further discussed later in Section 4.4.
    
\subsection{Training}
    In this section, we discuss the objective to optimize our adversarial network and one-stage recognition network. Let $I_i^{LR}$, $I_i^{HR}$ and $I_i^{SR}$ denote a low-resolution image, high-resolution image and SR image, respectively. Given a training dataset $\{I_i^{LR}, I_i^{HR}, text_{j=1}, ..., text_{j=count_i}, count_i\}_{i=1}^{N}$, our goal is to learn the adversarial model that predicts SR image from low-resolution image and recognition model that predicts character's class and location from SR image.
    
    \textbf{Pixel-wise loss} To force the generated plate image to high-resolution ground truth, our generator network is optimized for the MSE loss in each pixel values between the generated image sets and the small and blurry plate image sets calculated as follows:
    	\begin{equation}
    	    \begin{split}
        	    L_{MSE} = &\frac{1}{N}\sum_{i=1}^{N}(\Vert{G_{S1}}(I_i^{LR} - I_i^{HR})\Vert^2 \\
        	    & + \Vert G_{S2}({G_{S1}}(I_i^{LR}) - I_i^{HR})\Vert^2),
    	    \end{split}
    	\end{equation}
    where $G_{{S1}}$ means HR generator, $G_{{S2}}$ denotes the reconstruction network, and $w$ are the parameters of generator network.
    
    \textbf{Adversarial loss} In order to provide a sharpening effect to the generated image different from the MSE loss that gives the smoothing effect, we define adversarial loss as:
    	\begin{equation}
        	L_{adv} = { \frac{1}{N}\sum_{i=1}^N(\log(1 - D_{\theta}(G_w(I_i^{LR}))) + \log(D_{\theta}(I^{HR})))}
    	\end{equation}
    Adversarial loss amplifies the photo-realistic effect and is trained in the direction of deception of the discriminator.
    
    \textbf{Reconstruction loss} In order to let the quality of generated images by the $S1$ to be more photo-realistic, we propose the reconstruction loss that corrects changes in the generated image topology that interfere with the detection and is defined as follows:
    	\begin{equation}
        	L_{const} = \frac{1}{N}\sum_{i=1}^N{\Vert{G_{_{S1}}(I_i^{LR})} - G_{_{S2}}(G_{_{S1}}(I_i^{LR})) \Vert}.
    	\end{equation}
    The reconstruction loss is calculated as L1 loss, the difference between the output of $G_{S1}$ and $G_{S2}$.
    
    \textbf{Classification loss} The classification loss is playing both the roles of an character counting task as well as the discrimination task. To be more specific, the discriminator takes an image as input and classified it into two outputs: the HR real natural image or the SR fake image and the numbers of characters respectively. The loss of this multi-task is calculated as follows:
    	\begin{equation}
    	\begin{split}
        	L_{clc} = &\frac{1}{N}\sum_{i=1}^N(\log((y_i \wedge count_i) - D_{\theta}(G_w(I_i^{LR}))) \\
        	&+ \log((y_i \wedge count_i) - D_{\theta}(I_i^{HR}))),
    	\end{split}
    	\end{equation}
    where $y_n$ represents prediction value of the number of characters and the operations $\wedge$ with $y_n$ and $count_i$ output 1 if it predicts correctly or 0 respectively.

\section{\uppercase{Experimental results}}

\subsection{Setup}
     All the reported implementations are based on TensorFlow as learning framework, and our method has done on the NVIDIA TITAN X GPU. First of all, we use the YOLO-v3 for the pre-trained model on COCO \cite{lin2014microsoft} as our one-stage recognition model so that we trained license plate images by fine-tuning their network parameters. 
     
     Also, to avoid the premature convergence of the discriminator network, the generator network is updated more frequently than original one. In addition, higher learning rate is applied to the training of the generator. For stable training, we use a technique called gradient clipping trick \cite{pascanu2013difficulty} and the Adam optimizer \cite{kingma2014adam} with a high momentum term. For the discriminator network, we use the VGG-19 \cite{simonyan2014very} model pre-trained on ImageNet as our backbone network and we divide all the fully connected layers into two parallel $f_{count}$ and $f_{GAN}$ layers. The weights in all parallel fully connected layers are initialized from the standard Gaussian distribution with zero-mean, a standard deviation of $0.01$ and the constant $0$ as the bias in all layers. All models are trained on loss function for first 10 epochs with initial learning rate of $10^{-4}$. After that, we set the learning rate to a further reduced $10^{-5}$ for the remaining epochs. Finally, batch normalization \cite{ioffe2015batch} is used in all layers of generator and discriminator, except the last layer of the $G$ and the first layer of the $D$. 
\subsection{Dataset}

    \begin{figure*}[t]
    \begin{center}
       \includegraphics[width=1.0\linewidth]{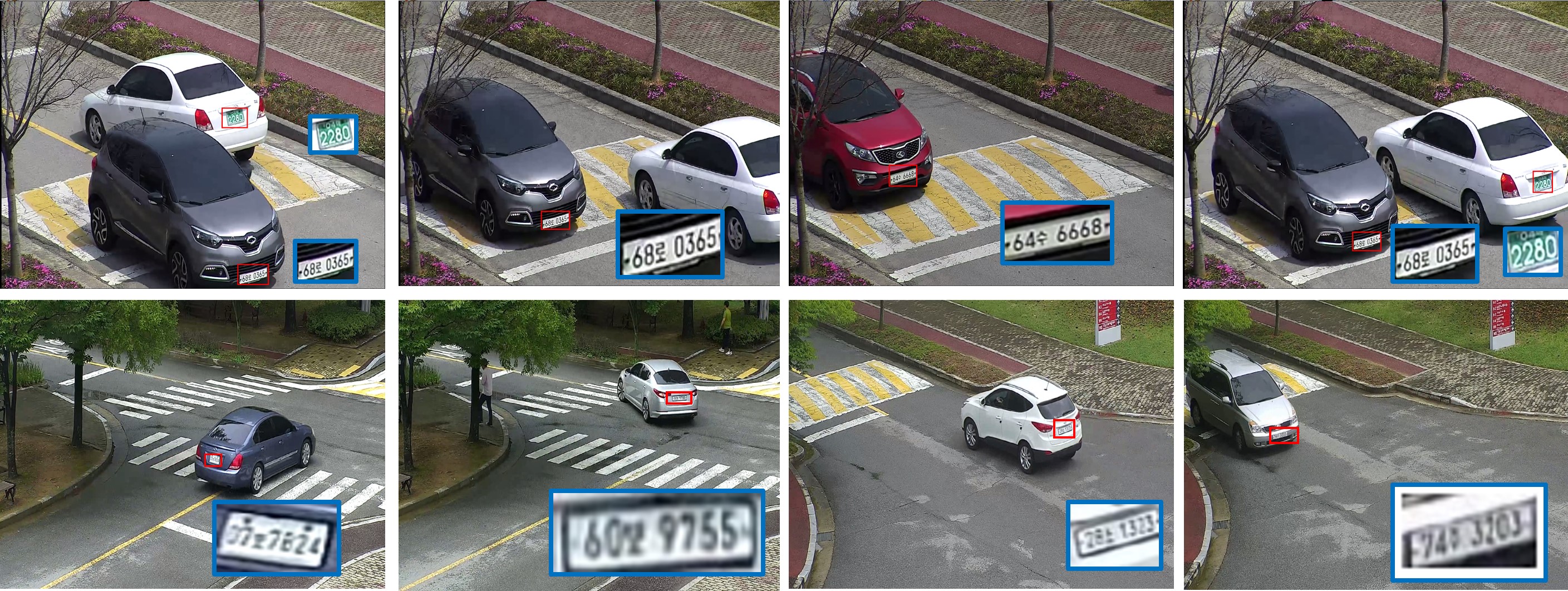}
    \end{center}
       \caption{Samples from the unconstrained surveillance cameras in GIST-LP dataset.}
    \label{fig:long}
    \label{fig:onecol}
    \end{figure*}
    
    \textbf{AOLP} : This dataset\cite{hsu2013application} includes 2,049 images of Taiwan license plates, which are collected from the unconstrained surveillance scenes. AOLP dataset is divided into three subsets: access control (AC) with 681 samples, traffic law enforcement (LE) with 757 samples, and road patrol (RP) with 611 samples, based on diverse application parameters. 100 samples per subset are used for the training, and the rest of the 581(AC)/657(LE)/511(RP) samples are used for testing. More specifically, AC has a narrow range of variation conditions, while LE/RP have a wider range of variation conditions. Therefore, compared to the AC subset, LE/RP are more challenging subsets because they require a wider range of search conditions on the experiments. Besides, the RP samples collected via mobile have more challenging conditions because of the larger pan and orientation changes compared to the LE samples collected at road cameras with fixed viewing angles. 
  
    \textbf{GIST-LP} : We collected and annotated a new dataset GIST-LP for LPR. Our dataset is targeted on images captured from surveillance cameras under unconstrained scenes. We do not limit the license plate always to be large and front. We used traffic surveillance cameras which has 1920 x 1080 pixels of spatial resolution. We annotated the characters, including Korean (30 categories) and numbers (0-9, 10 categories) for all of the license plate images. In total, there are 1,800 license plates that appear in 1,569 frames. For license plate images, the characters are usually small-sized, blurred or tilted without occlusion. The dataset include information about bounding box for each character and text class (Koreans and numbers).

    \setlength{\tabcolsep}{10pt}
    \begin{table*}
    \begin{center}
    \caption{Comparison of our method with other state-of-the-art method on the AOLP dataset. 
    }
    \label{table:headings}
    \begin{tabular}{l|ccc | c}
    \hline\hline
    {Method}       & \multicolumn{4}{c}{Performance}                          \\ \cline{2-5} 
                                  & AC         & LE         & \multicolumn{1}{l|}{RP}         & Avg     \\ \hline\hline
    \cite{anagnostopoulos2006license}          & 92.00\%          & 88.00\%          & \multicolumn{1}{c|}{91.00\%}          & 86.34\%          \\
    
    \cite{jiao2009configurable} & 90.00\%          & 86.00\%          & \multicolumn{1}{c|}{90.00\%}          & 88.51\%           \\
    
    \cite{smith2007overview}             & 96.00\%          & 83.00\%          & \multicolumn{1}{c|}{83.00\%}          & 87.31\%           \\
    
    \cite{hsu2013application}        & 95.00\%          & 93.00\%        & \multicolumn{1}{c|}{94.00\%}          & 94.17\%          \\\hline\hline
    Baseline (YOLO v3) \cite{redmon2018yolov3}                     & 94.66\%          & 89.04\%          & \multicolumn{1}{c|}{89.04\%}          & 90.90\%         \\ 
    without pixel-wise MSE loss                     &97.24\%          & 94.67\%          & \multicolumn{1}{c|}{94.91\%}          & 95.60\%         \\ 
    without reconstruction loss                     & 96.21\%          & 88.89\%          & \multicolumn{1}{c|}{94.32\%}          & 92.91\%         \\ 
    without adversarial loss                     & 95.18\%          & 87.67\%          & \multicolumn{1}{c|}{93.93\%}          & 92.00\%         \\ 
    without classification loss                     & 96.39\%          & 94.98\%          & \multicolumn{1}{c|}{96.48\%}          & 95.88\%         \\ 
    \textbf{Ours}                          & \textbf{97.59\%} & \textbf{95.89\%} & \multicolumn{1}{c|}{\textbf{96.87\%}} & \textbf{96.74\%} \\ \hline\hline
    \end{tabular}
    \end{center}
    \end{table*}
    \setlength{\tabcolsep}{1.4pt}
    
    \setlength{\tabcolsep}{10pt}
    \begin{table*}
    \begin{center}
    \caption{Comparison of our method with other state-of-the-art method on the GIST-LP dataset.}
    \label{table:headings}
    \begin{tabular}{l|c}
    \hline
    Method $\qquad\qquad\qquad$& Performance \\
    \hline\hline
    \noalign{\smallskip}
    RCNN based on VGG-16 \cite{girshick2014rich} & 74.44\% \\ 
    RCNN based on ZFNET \cite{girshick2014rich} & 72.11\% \\  
    Faster-RCNN \textit{et al.} \cite{ren2015faster} & 86.77\% \\ \hline\hline 
    Baseline (YOLO v3) \cite{redmon2018yolov3} & 84.16\% \\ 
    Ours without pixel-wise MSE loss & 91.78\% \\ 
    Ours without reconstruction loss & 89.00\% \\ 
    Ours without adversarial loss & 87.72\% \\ 
    Ours without classification loss & 90.78\% \\ 
    \textbf{Ours} & \textbf{93.83\%} \\ 
    \hline
    \end{tabular}
    \end{center}
    \end{table*}
    \setlength{\tabcolsep}{1.4pt} 
    
    \begin{figure}[t]
    \begin{center}
       \includegraphics[width=1.0\linewidth]{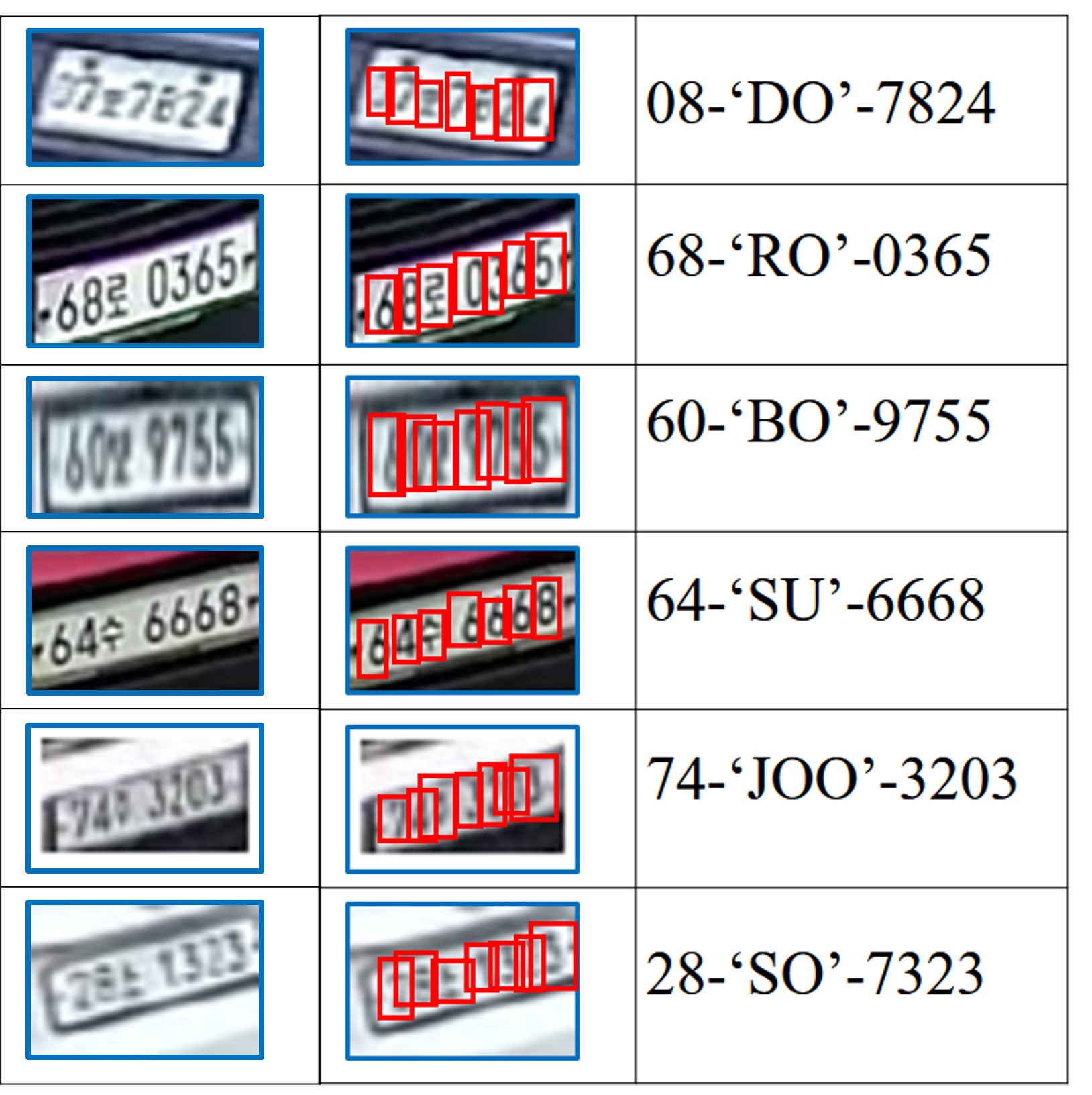}
    \end{center}
       \caption{Example in GIST-LP dataset \cite{laroca2018robust}. Qualitative sample images of recognition results. The first column shows the original plates, the second column shows the character localization results and the third indicates the recognitionm results.}
    \label{fig:long}
    \label{fig:onecol}
    \end{figure}

\subsection{Comparison with Other Methods}
    
    \begin{figure*}[t]
    \begin{center}
       \includegraphics[width=1.0\linewidth]{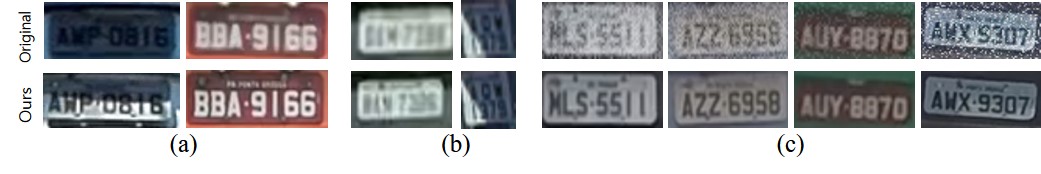}
    \end{center}
       \caption{Example in AOLP dataset \cite{hsu2013application}. Poor-resolution and background clutter are common challenging issues on character recognition problem.}
    \label{fig:long}
    \label{fig:onecol}
    \end{figure*}
    
    In the experiment with AOLP, we compared our method with the state-of-the-are license plate recognition approaches \cite{anagnostopoulos2006license,jiao2009configurable,smith2007overview,hsu2013application}. The results are listed in Table 1, which are provided with accuracy of recognition to evaluate both text localization and classification are all performed well at the same time. We see that our method obtained the highest performance (\textit{i.e.} 96.74\%) on the all subsets, and outperformed the state-of-the-are LPR approaches by more than 2.5\%. Also, it is important to note that, under the fairly tilted conditions, our method operated consistently robust and successfully detects the characters, while the baseline fail to detect. Furthermore, one interesting finding of these results is that, based on Figure 6 (b,c), the addition of adversarial loss lead to the highlighting of the positive features, while decimating of other irrelevant features. By doing so, it was further improved when detecting under night or confusing conditions. Based on these observations, our proposed method operated at least as well as others, which outperformed all other methods in most cases.
    
    To show the results of experiment of LPR with GIST-LP, we compared our method with \cite{girshick2014rich,ren2015faster} and followed the standard metrics (\textit{i.e.} accuracy of recognition) of the GIST-LP. There were many tiny license plates in GIST-LP, making character detection not be accurate. Hence, we found that the state-of-the-art method \cite{redmon2018yolov3} that performed without considering the tiny size and blurred condition recorded on the inferior performance. However, our method mitigated the influence of these conditions and indicated these license plates successfully. Under such a challenging condition, our LPR performance still achieved a comparable performance (93.83\%) over all other state-of-the-art LPR approaches, as shown Table 2.

\subsection{Ablation Study}
    In the proposed method, the loss functions of adversarial networks locate different regions, each with their unique roles. In order to inspect its influence on character recognition performance, we removed one loss function from the objective function at a time and performed an ablation study with it to compare the complete objective function. Most extremely, we perform experiments that compare the baseline and overall objective function, which obtain the superior performance by a considerable gap (5.84\% / 9.67\%) from Table 1 and 2. 
    
    Also, when removing one loss function from the overall objective function our method shows a considerable performance drop. First of all, even if the MSE loss is not suitable for tiny objects due to the smoothing effect, if there is no MSE loss, the performance degradation is up to 1.14\% (in AOLP) / 2.05\% (in GIST-LP), affecting the image up-scaling super-resolution. Then the reconstruction loss affects the correct converting of the tilted plate, because the SR performance of the generator is somewhat dependent on the degree of tilted angle of the license plate, and it leads to about 3.83\% (in AOLP), 4.83\%(in GIST-LP) improvement in performance. In another step, we observe that adversarial loss leads to the sharpened super-resolved result of minimax game. Thus it has a great influence on the detection performance as shown in Figure 6. The GIST-LP dataset which has relatively more tiny plates than AOLP dataset has found a performance improvement of almost 4.74\% as shown Table 2, and the AOLP dataset also achieves performance improvement of nearly 6.11\% as shown Table 1. Finally removing classification loss in the objective function shows a significant impact on the character recognition performance, which observes an impressive improvement of 0.86\% (in AOLP) and 3.15\% (in GIST-LP). This proves that our two parallel fully-connected layers for classification affect the classification performance for our text localization of the detector as well as the SR performance of the generator. Also, we demonstrate that the counting term as conditional data benefits to better explore the space of the character localization as much as possible.
    
\subsection{Qualitative Results} 
As shown in Figure 6., we give additional examples of the clear LP generated by the proposed generator network from the tiny ones. Upon thorough investigation of the generated images, we find that our method learn strong priors using the proposed new loss functions of GAN by focusing on images of plate contour, certain letters and numbers as shown in Figure 6 (a). It implies that the proposed loss significantly allows visually clearer LP and can be used to solve the ill-posed problem. Thus, SR module can capture the tiny LP without hallucination and it implies the proposed architecture has an impact on reducing the false negatives.

\section{Conclusions}
\label{sec:conclusion}
    In this paper, we propose a new method based on GAN to recognize characters in unconstrained license plates. We design a novel network to directly generate a clear SR image from a blurry small one, and our up-sampling sub-network and reconstruction sub-network are trained in an end-to-end way. Moreover, we introduce an extra classification branch to the discriminator network, which can distinguish the HR/SR and the character counting probability simultaneously. Furthermore, the adversarial loss brings to generator network to restore a clearer SR image. Our experiment on AOLP and GIST-LP datasets demonstrate the substantial improvements, when compared to previous state-of-the-art methods.

\section*{\uppercase{Acknowledgements}}

\noindent This work was supported by institute for the information \& communications technology promotion (IITP) grant funded by the Korean government (MSIP) (B0101-16-0525, development of global multi-target tracking and event prediction techniques based on real-time large-scale video analysis.

\vfill
\bibliographystyle{apalike}
{\small
\bibliography{example}}

\vfill
\end{document}